\documentclass[conference]{IEEEtran}
\IEEEoverridecommandlockouts
\usepackage{cite}
\usepackage{amsmath,amssymb,amsfonts}
\usepackage{algorithmic}
\usepackage{graphicx}
\usepackage{textcomp}
\usepackage{xcolor}
\usepackage{eso-pic}
\usepackage{atbegshi}

\def\BibTeX{{\rm B\kern-.05em{\sc i\kern-.025em b}\kern-.08em
    T\kern-.1667em\lower.7ex\hbox{E}\kern-.125emX}}

\makeatletter

\def\ps@IEEEtitlepagestyle{%
  \def\@oddfoot{}%
  \def\@evenfoot{}%
}

\@ifundefined{showcaptionsetup}{}{%
  \PassOptionsToPackage{caption=false}{subfig}}
\usepackage{subfig}

\newcommand\PlaceHeaderLeft[1]{%
  \AtPageUpperLeft{%
    \put(\LenToUnit{1cm},\LenToUnit{-1cm}){%
      \parbox{\textwidth}{\fontsize{9}{11}\selectfont #1}%
    }%
  }%
}

\newcommand{\conf}[1]{%
  \AddToShipoutPictureBG*{%
    \PlaceHeaderLeft{#1}%
  }%
}

\makeatother

\begin{document}

\title{Pediatric Bone Age Prediction Using Deep Learning%
\thanks{Accepted at ICCIT 2023. This version is the author-prepared manuscript. The final published version appeared in IEEE Xplore.}}

\conf{2023 26th International Conference on Computer and Information Technology (ICCIT)\\
13--15 December 2023, Cox’s Bazar, Bangladesh}

\author{
\IEEEauthorblockN{
Al Zadid Sultan Bin Habib\textsuperscript{1},
Md. Ekramul Islam\textsuperscript{2},
Md Asif Bin Syed\textsuperscript{3},
Md Younus Ahamed\textsuperscript{4},
Tanpia Tasnim\textsuperscript{5}
}
\IEEEauthorblockA{
\textsuperscript{1,4}Lane Department of Computer Science and Electrical Engineering, West Virginia University, Morgantown, WV 26506, USA\\
\textsuperscript{2}Department of Computer Science \& Engineering, Stamford University Bangladesh, Dhaka-1217, Bangladesh\\
\textsuperscript{3}Department of Industrial and Management Systems Engineering, West Virginia University, Morgantown, WV 26506, USA\\
\textsuperscript{5}Department of Computer Science and Engineering, Green University of Bangladesh, Narayanganj-1461, Dhaka, Bangladesh\\
Email: ah00069@mix.wvu.edu\textsuperscript{1}, eislam706@gmail.com\textsuperscript{2}, ms00110@mix.wvu.edu\textsuperscript{3}, ma00087@mix.wvu.edu\textsuperscript{4}, tanpia@cse.green.edu.bd\textsuperscript{5}
}
}

\maketitle
\begin{abstract}
Pediatric bone age prediction is a crucial task in clinical practice that can help diagnose endocrine disorders and provide insight into a child's growth and development. However, conventional bone age prediction methods are often labor-intensive and require specialized radiological expertise. This paper presents a Deep Learning (DL)-based approach to pediatric bone age prediction using EfficientNet with Additive Attention, a state-of-the-art neural network architecture for image classification and regression tasks. The method utilizes over 12,000 X-ray images from the RSNA bone age dataset. It involves image preprocessing, transforming them into three-channel images, and training a Convolutional Neural Network (CNN) to automatically learn the features of hand bone images. This approach provides a more effective and accurate solution for predicting bone age, which is critical in diagnosing pediatric endocrine diseases. This work uses two variations of the EfficientNet model (B0 and B4), where EfficientNetB4 is also finetuned with the Additive Attention mechanism. These three models predict the age for the original age, and their comparison is shown in curves. The predicted ages depict that in most cases, EfficientNetB4 and EfficientNetB4 with Additive Attention (EN-AA) successfully predicted the bone ages more accurately regarding the original age, and their performance was better than the EfficientNetB0. Specific performance metrics are provided to underscore this improvement. Learning curves for training and validation loss confirm effective learning without overfitting or underfitting, further validating our approach's efficacy in pediatric endocrine disease diagnosis.
\end{abstract}


\section{Introduction}
In the rapidly evolving fields of Computer Science (CS), technology, and medical science, numerous cross-disciplinary advancements have significantly impacted healthcare. Traditionally, medical procedures like X-ray photo scanning and analysis were performed manually. However, with the advent of computer image processing technologies, these tasks have become more efficient and accurate. In medical practice, age is a crucial measure of human growth, but bone age is often a more accurate indicator of biological maturity, as detailed in \cite{b12}. Typically, an X-ray of the left wrist is used to assess bone age, which is essential in evaluating adolescent development, screening for genetic disorders, and talent assessment. Traditionally, most X-ray analyses involve manual comparison of a patient's radiograph with an age-specific atlas or using a bone-specific scoring system. However, these manual methods are time-consuming, subject to limitations, and prone to errors.\\

Numerous domestic and international initiatives have emerged focusing on X-ray bone age image learning, reflecting the growing intersection of medical imaging and CS. Advanced software solutions such as BoneXpert \cite{b13} have employed Computer Vision (CV) techniques to reconstruct hand bone contours. However, challenges still need to be solved, particularly with low-quality images. The advent of Artificial Intelligence (AI) technology, alongside advancements in computer and graphics technology, offers new avenues for tackling these interdisciplinary challenges. The field of bone age assessment has progressively shifted from manual to automated evaluations, thanks to AI and DL innovations. In this context, our work utilizes the variations of the EfficientNet model with the Additive Attention mechanism\cite{b1}, showcasing its application in automated and precise bone age assessment.

This network optimizes the network depth, width, and image resolution balance to achieve ideal performance. EfficientNet surpasses traditional models in size-to-accuracy ratio \cite{b14}. A separable CNN standardizes and normalizes the input for processing X-ray images. This involves segmenting the hand area for image registration, focusing on the central region of interest, and rotating key points to isolate the hand \cite{b15}. Utilizing the DeepLabv3 plus \cite{b2} and MobileNetV1 \cite{b3} architectures, the network employs separable convolutions as its core unit. We apply various rotations, scaling, and cropping techniques to enhance the dataset and prevent underfitting. Previous research has yet to extensively explore bone age estimation in hand X-ray images using the EfficientNet model. We aim to establish a rapid and efficient method for this task, leveraging the latest advancements in DL models. For further details, readers are directed to \cite{b3,b4,b5,b6,b7,b8,b9} and related references.

This study leverages DL and image processing techniques for predicting bone age from over 12,000 children's hand bone X-ray images in the RSNA bone age dataset \cite{b10}. The approach includes three key steps: (1) Preprocessing the X-ray images: size normalization, noise reduction, and histogram equalization, to standardize the images for DL. (2) Feature extraction using the EfficientNet model, focusing on extracting relevant features from hand bone images. (3) Applying a CNN for DL to automatically extract features and accurately determine bone age.

The key contributions of this work are summarized as follows:

\begin{itemize}
\item[$\blacksquare$] Investigated the application of pre-trained EfficientNet models in the context of bone age prediction, demonstrating their effectiveness in a medical imaging domain.
\item[$\blacksquare$] Conducted a comparative analysis of various EfficientNet model variants, providing insights into their performance in bone age estimation, a critical task in pediatric healthcare.
\item[$\blacksquare$] Innovatively adapted the Additive Attention mechanism, traditionally used in NLP, to the EfficientNetB4 model, creating an enhanced EN-AA model for the specific demands of image regression tasks in medical imaging.
\end{itemize}

The subsequent sections of the paper are organized in the following manner: Section II outlines the methodological framework utilized in this work. At the same time, Section III offers a comprehensive explanation of the obtained results. The paper is concluded in Section IV by outlining the limitations and future research directions.
\section{Methodological Framework}

\subsection{Preprocessing}
\begin{figure}
	\centerline{\includegraphics[width=90mm, height=60mm]{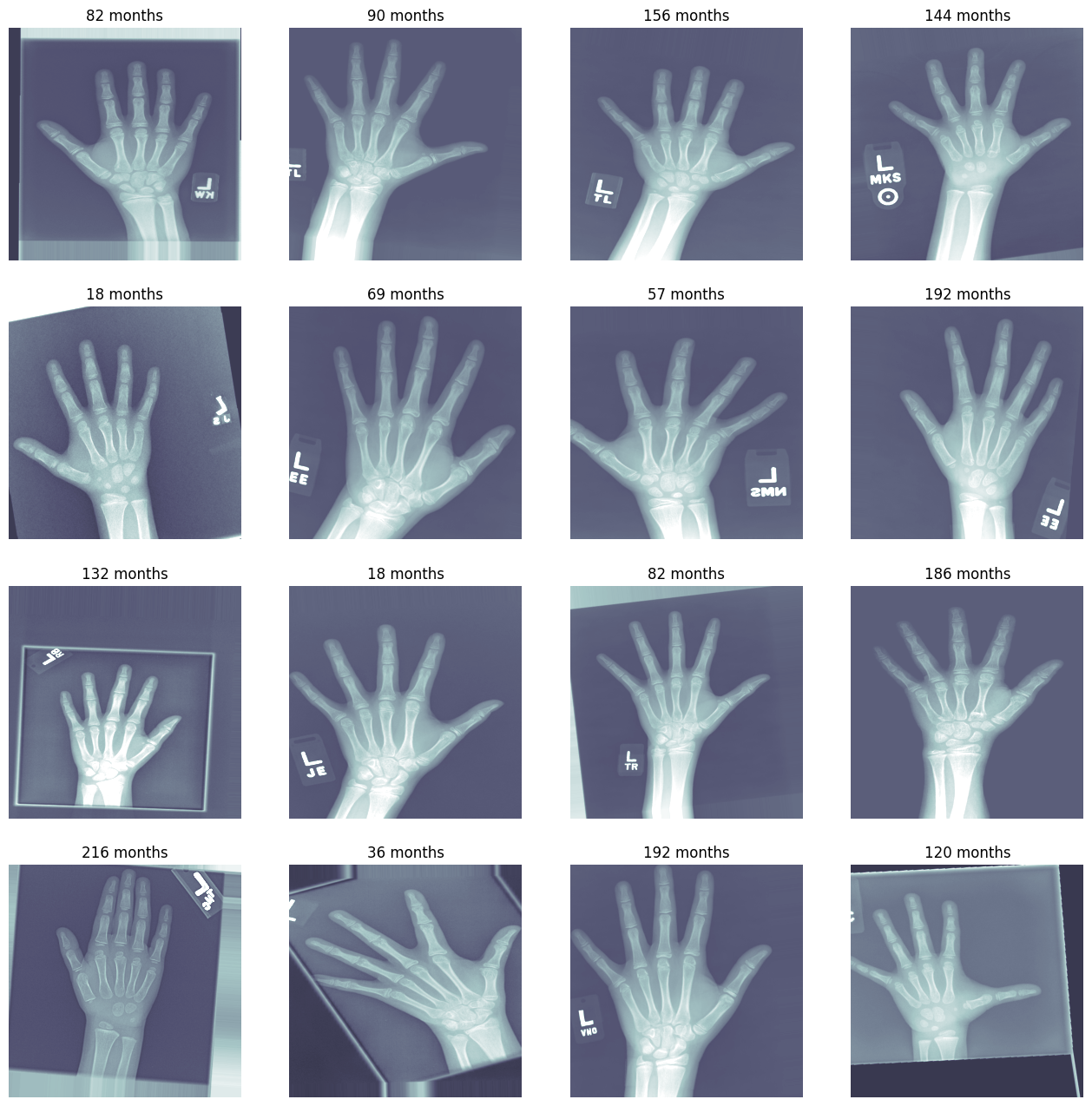}}
	\caption{Sample X-ray bone images.}
	\label{fig:1a}
\end{figure}

The dataset for our study, sourced from RSNA 2017 \cite{b10}, consists of over 12,000 X-ray images showcasing children's hand bones, including a sample depicted in Fig: \ref{fig:1a}. Accompanying each image are relevant details like age and gender. A notable characteristic of this dataset is the variation in image resolution and size, with the dimensions of hand bone images generally ranging from 800 to 2200 pixels in length and width, as identified through statistical analysis. The RSNA X-ray images were divided into two sets for model training and evaluation: 70\% for training and validation and 30\% reserved for testing.

\begin{figure}
\centering
\includegraphics[width=0.5\textwidth]{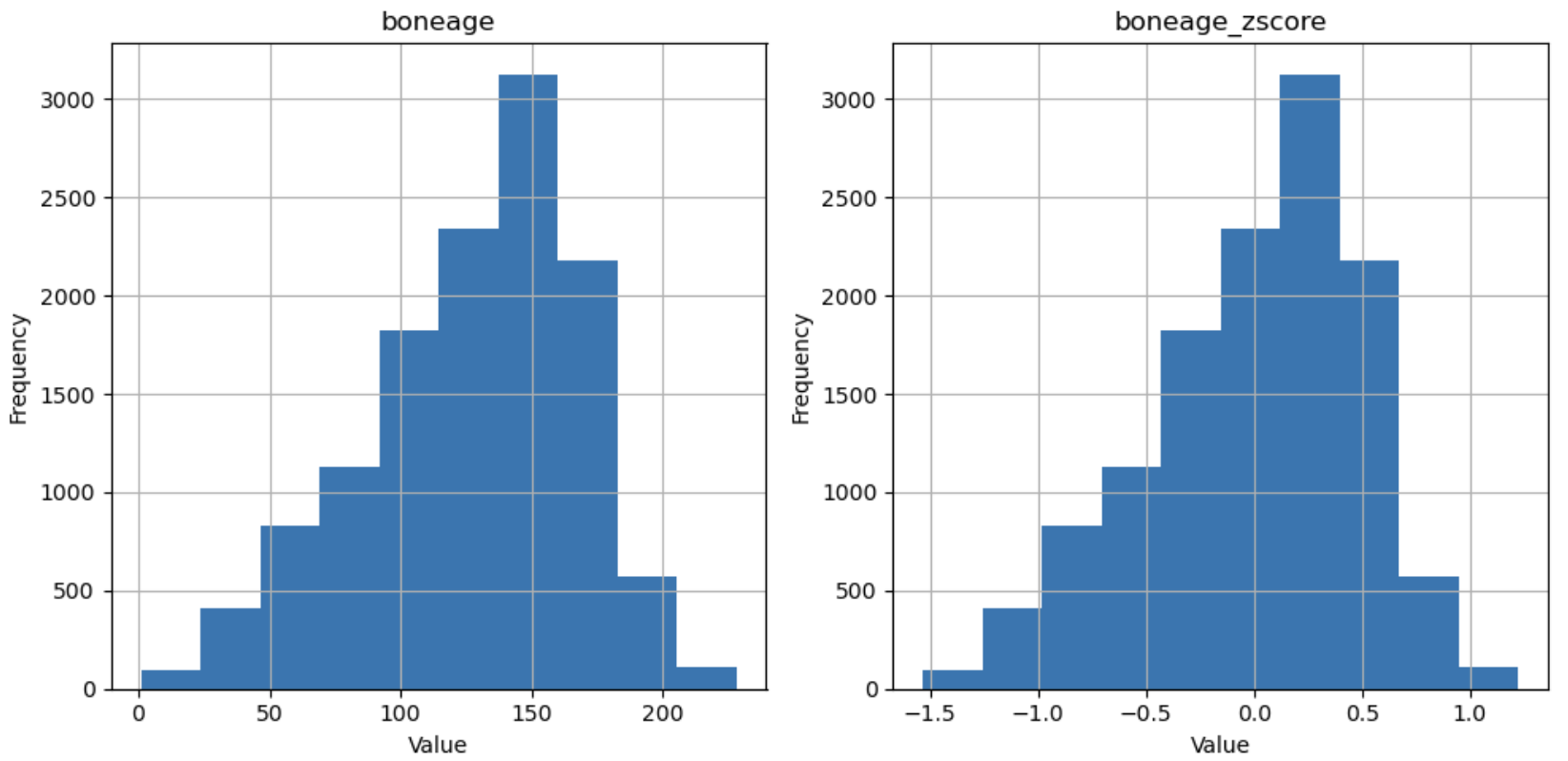}
\caption{Histogram of the bone ages before and after normalization.} 
\label{fig:1b}
\end{figure}

\subsection{EfficientNet Model}
EfficientNetB0, the most compact model within the EfficientNet family \cite{b1}, distinguishes itself with only 5.3 million parameters, far fewer than larger models like ResNet50 or InceptionV3. This efficiency makes it ideal for computational resource conservation. The architecture comprises seven blocks, each integrating convolutions, batch normalization, and activation functions, designed for an input size of 224x224 pixels, aligning with the ImageNet dataset standards. A notable feature of EfficientNetB0 is its compound scaling technique, balancing depth, width, and resolution, enhancing its performance in diverse CV applications, including image classification, object detection, and segmentation. Regarding effectiveness, EfficientNetB0 attains a top-1 accuracy of 76.3\%  and a top-5 accuracy of 93.2\% on the ImageNet dataset \cite{b18}, making it competitive among small-scale models.

\begin{figure}
	\centering
	\includegraphics[width=0.5\textwidth]{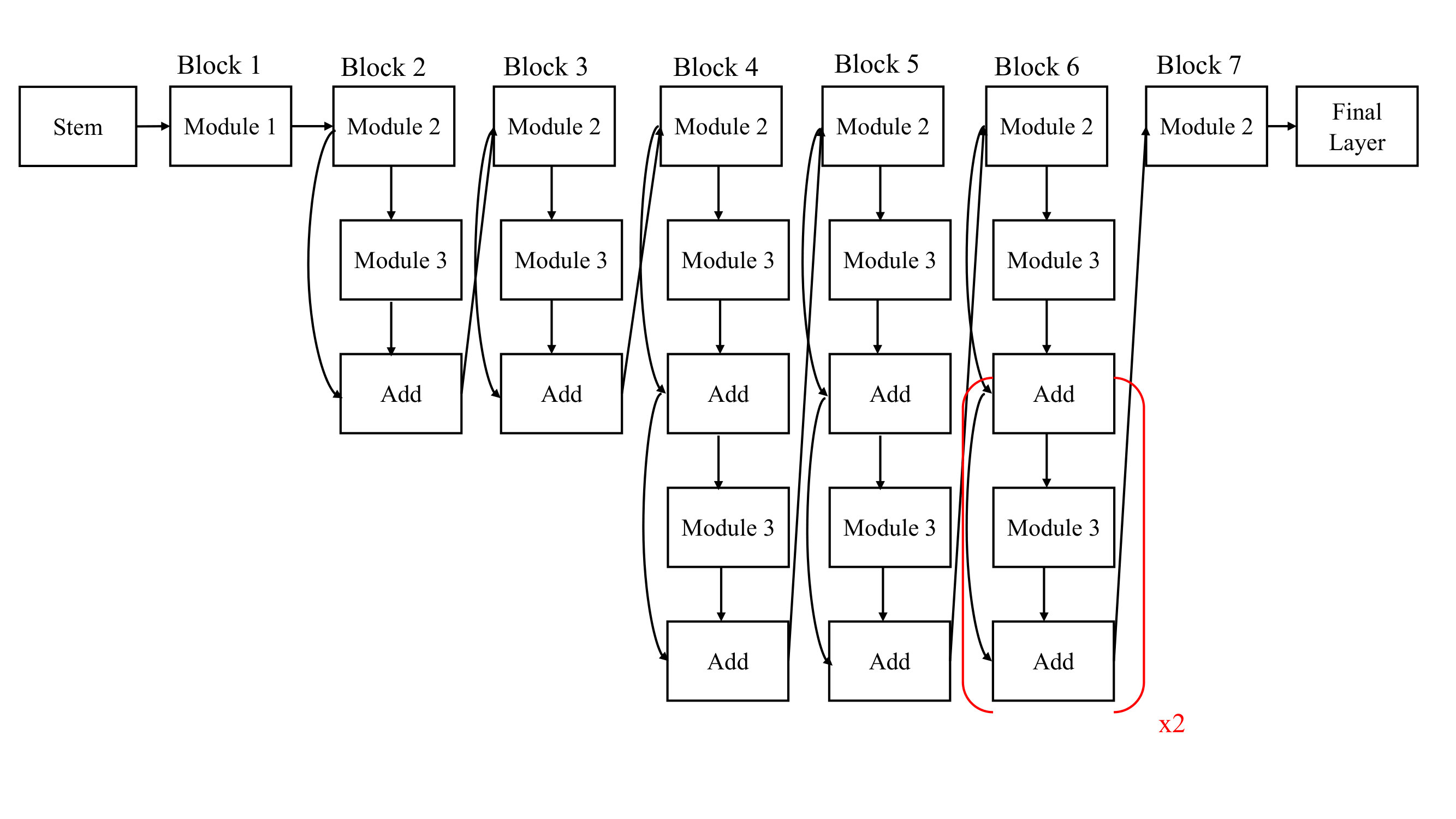}
	\caption{EfficientNetB0 architecture.} 
	\label{fig:2a}
\end{figure}

 EfficientNetB0, part of the EfficientNet series, is optimized for CV tasks using a modular design that combines multiple building blocks and subblocks. Building blocks feature convolutional layers with a bottleneck structure for efficiency, followed by Squeeze-and-Excitation (SE) modules for feature recalibration. Subblocks, which are more straightforward and used in early layers, lack the SE module and have fewer filters. The models incorporate a compound scaling method, adjusting filters, depth, and resolution for balanced architectural complexity. This design approach, coupled with a standardized stem structure comprising convolutional layers, batch normalization, and max pooling, enables EfficientNetB0 to process low-level features efficiently before advancing to complex tasks. Its stem's consistent design facilitates scalability and comparative analysis within the EfficientNet family, maintaining both efficiency and effectiveness.

\begin{figure}
	\centering
	\includegraphics[width=0.5\textwidth]{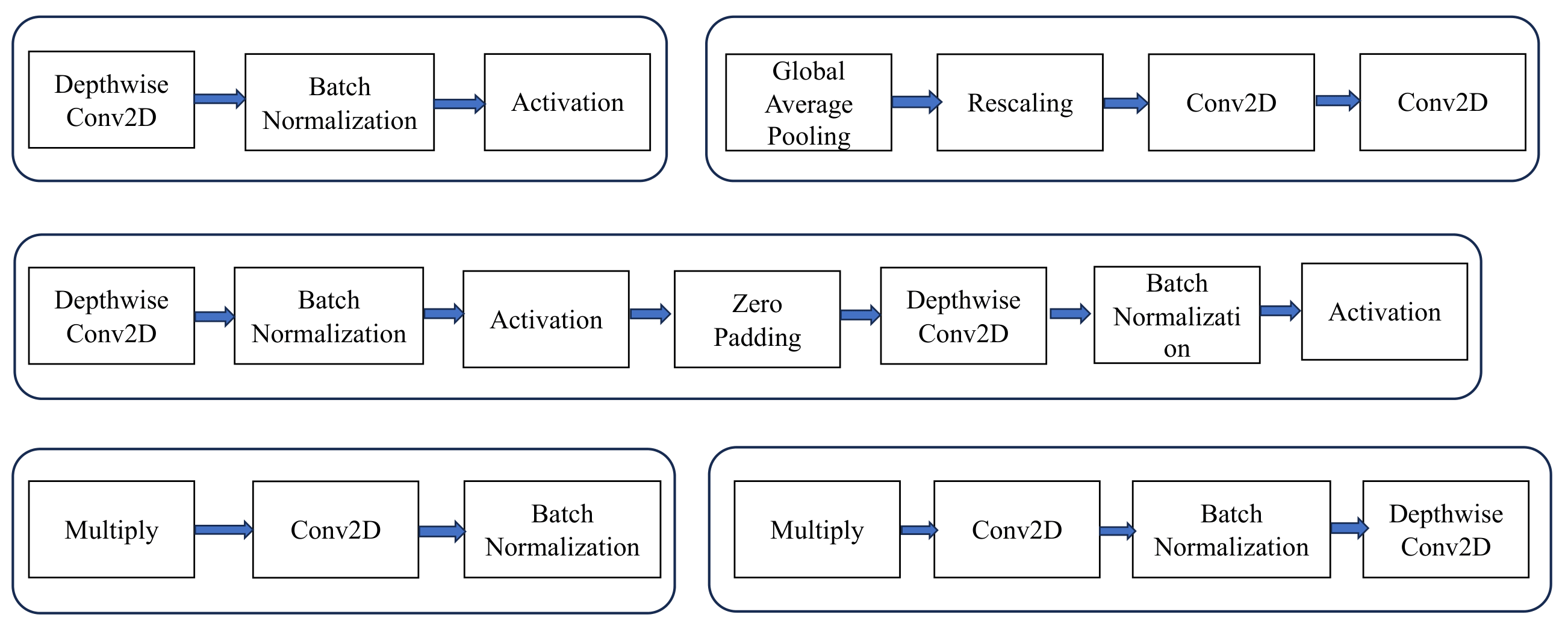}
	\caption{Modules in EfficientNet architectures.} 
	\label{fig:2c}
\end{figure}

\begin{figure}
	\centering
	\includegraphics[width=0.5\textwidth]{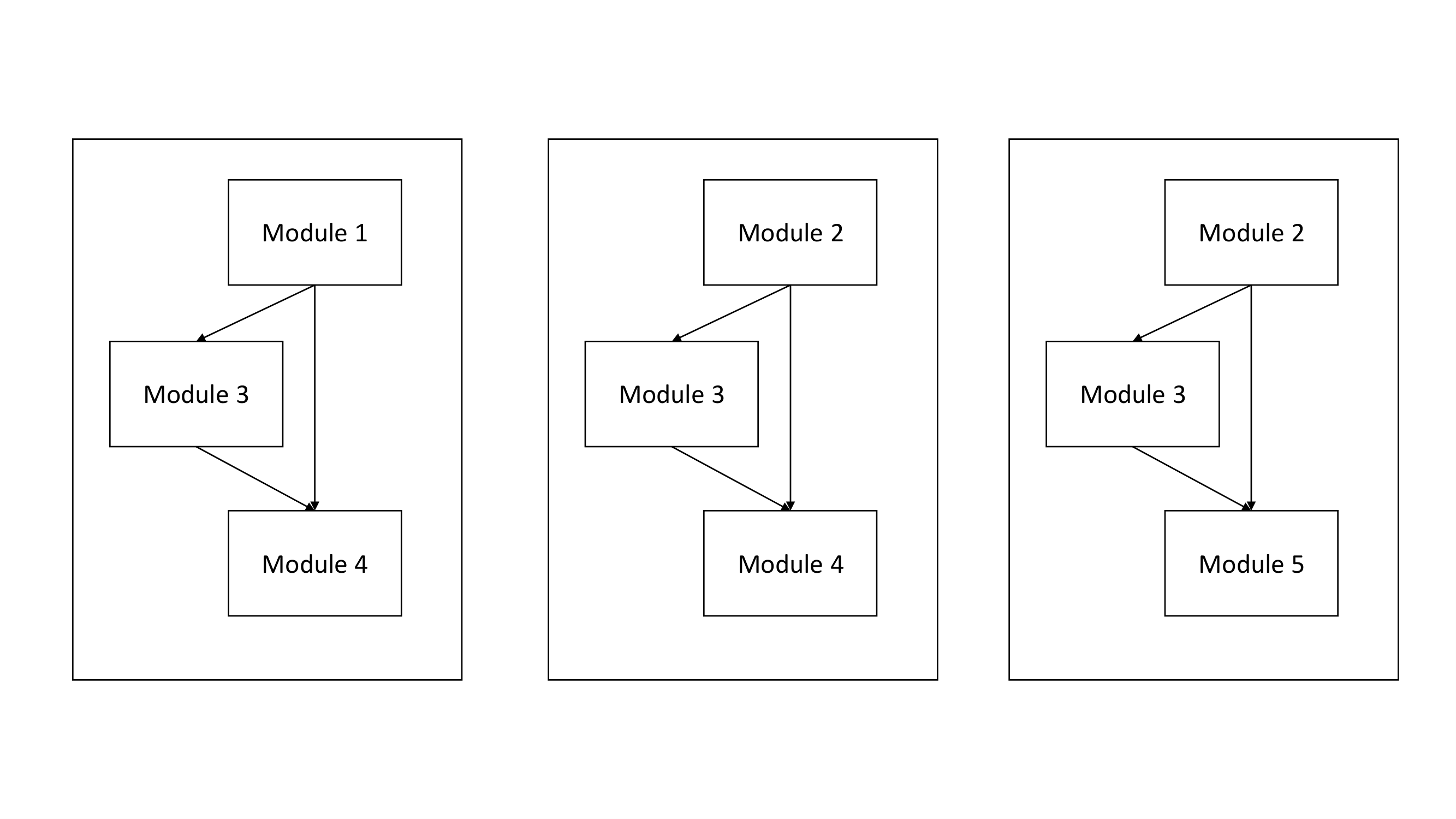}
	\caption{Sub-blocks in EfficientNet architectures.} 
	\label{fig:2d}
\end{figure}

\begin{figure}
\centering
\includegraphics[width=0.6\textwidth]{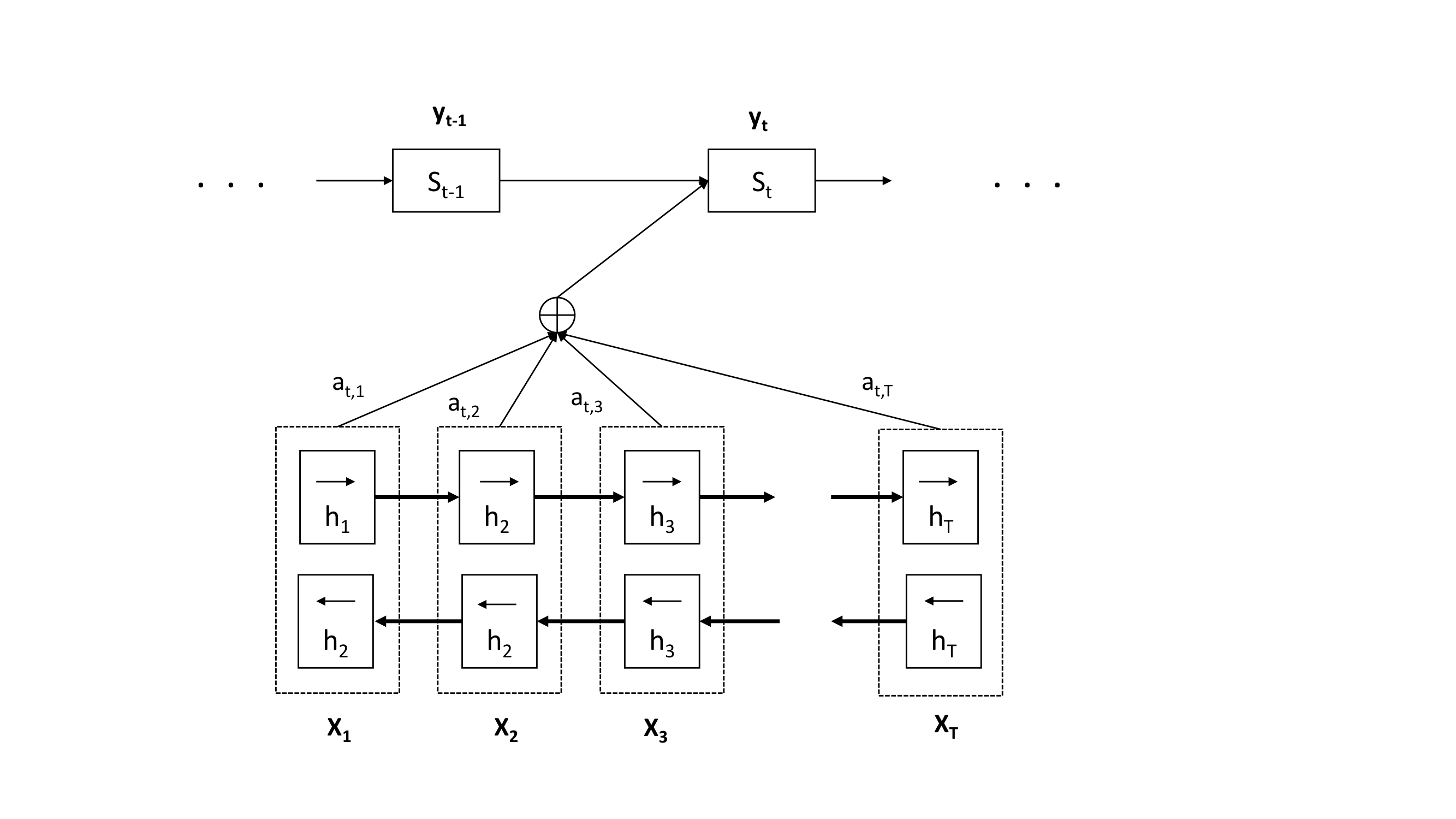}
\caption{Additive Attention architecture.} 
\label{fig:2e}
\end{figure}

\subsection{EN-AA}
\subsubsection{Image Model}
The initial segment of the model employs an EfficientNetB4, pre-trained on ImageNet, for extracting image features. It processes an input tensor with dimensions (256, 256, 1), outputting a tensor sized (8, 8, 1792) after the concluding convolutional layer. This tensor undergoes flattening through a flattened layer, resulting in a vector of dimensions (114688). Subsequently, this vector is fed into a sequence of three dense layers containing 128, 64, and 32 neurons, respectively, and employing the ReLU activation function. The final output from this image model is a tensor with dimensions (32).
\begin{figure}[htbp]
	\centering
	\includegraphics[width=0.5\textwidth]{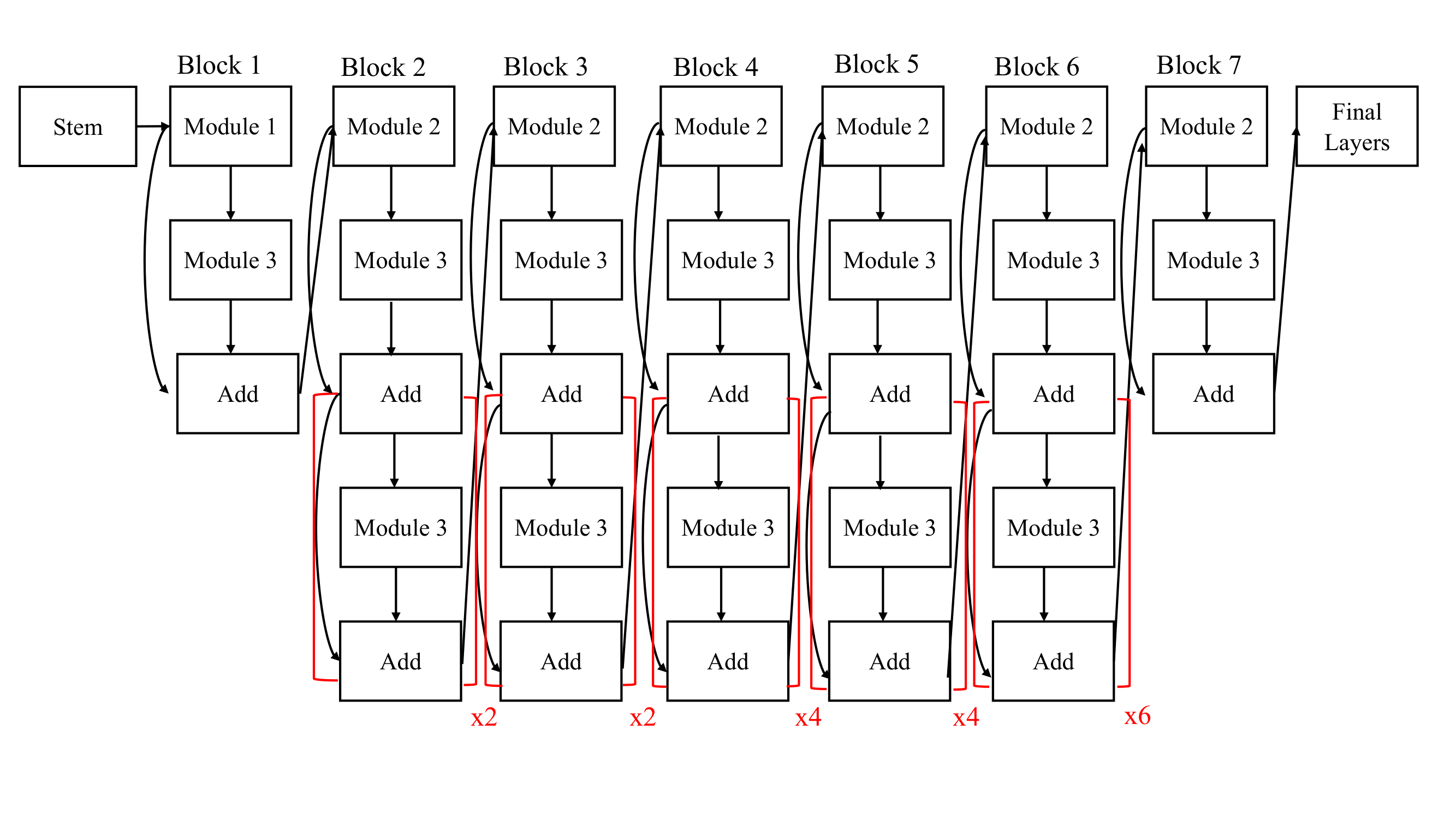}
	\caption{EfficientNetB4 architecture.} 
	\label{fig:3a}
\end{figure}
\subsubsection{Gender Model}
The second component of the model is dedicated to processing gender data through a straightforward dense network. It begins with an input tensor sized (1) and channels it through two dense layers. The first layer comprises 64 neurons and the second 32, utilizing the ReLU activation function. The output from this gender-focused model is a tensor with dimensions (32).
\subsubsection{Concatenate}
The output tensors from both the image and gender models are merged using the Concatenate layer in Keras. This operation forms a combined tensor with a shape of (64), effectively integrating the distinct feature sets derived from each model into a single, unified tensor for subsequent processing steps.
\subsubsection{Attention Mechanism}
The concatenated tensor is processed through an Additive Attention mechanism employing the previously defined additive attention function. The resultant output from this attention mechanism is a weighted sum of the concatenated tensor, effectively focusing on specific features within it \cite{b11}. This approach allows the model to selectively emphasize essential elements in the tensor, enhancing its overall interpretative capability.
\subsubsection{Dense Layers}
The weighted sum tensor undergoes further processing by passing through two dense layers. The first layer contains 32 neurons, and the second comprises 16, utilizing the ReLU activation function. This arrangement allows for additional refinement and extraction of features from the tensor.
\subsubsection{Output Layer}
The model's final output is generated by a single dense layer of one neuron equipped with a linear activation function. This layer predicts the target variable, translating the processed features into the outcome.
\subsubsection{Model Definition}
The model is built using Keras' Model class, combining image and gender input layers into a single output layer. This structure effectively merges distinct data sources for a holistic prediction.
\section{Results Analysis}
\label{res}
\begin{figure}
	\centering
	\includegraphics[width=0.5\textwidth]{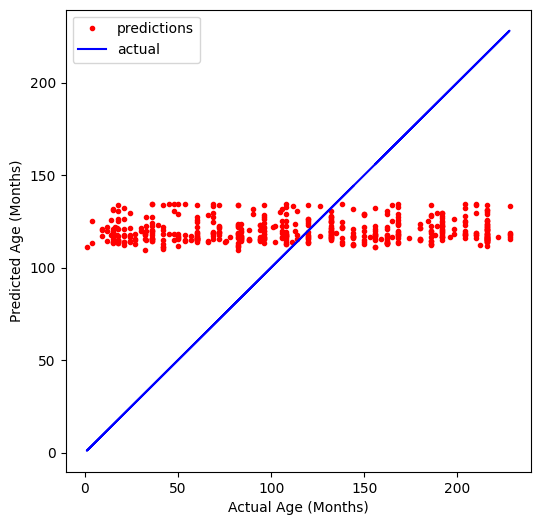}
	\caption{Predicted age by EfficientNetB0.} 
	\label{fig:4a}
\end{figure}

\begin{figure}
	\centering
	\includegraphics[width=0.5\textwidth]{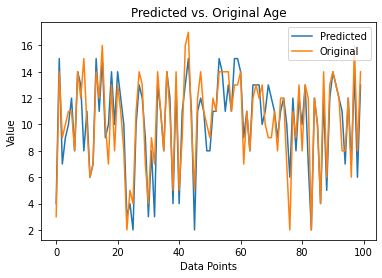}
	\caption{Predicted age by EfficientNetB4.} 
	\label{fig:4e}
\end{figure}

\begin{figure}
	\centering
	\includegraphics[width=0.5\textwidth]{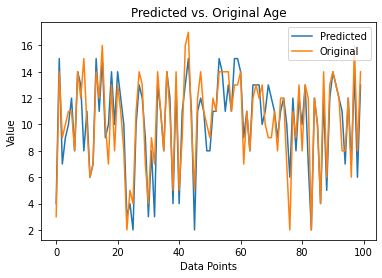}
	\caption{Predicted age by EN-AA.} 
	\label{fig:4f}
\end{figure}

\begin{figure}
	\centering
	\includegraphics[width=0.5\textwidth]{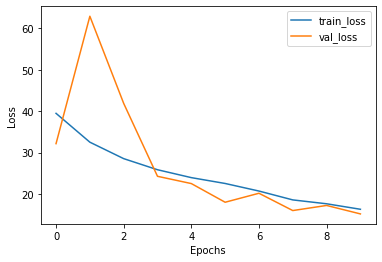}
	\caption{Learning curves for EfficientNetB4.} 
	\label{fig:4g}
\end{figure}

\begin{figure}
	\centering
	\includegraphics[width=0.5\textwidth]{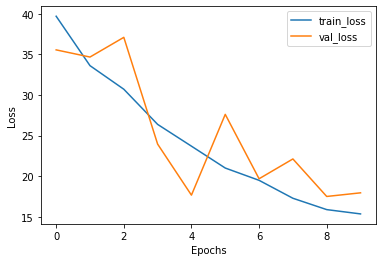}
	\caption{Learning curves for EN-AA.} 
	\label{fig:4h}
\end{figure}

\begin{figure}
	\centering
	\includegraphics[width=0.5\textwidth]{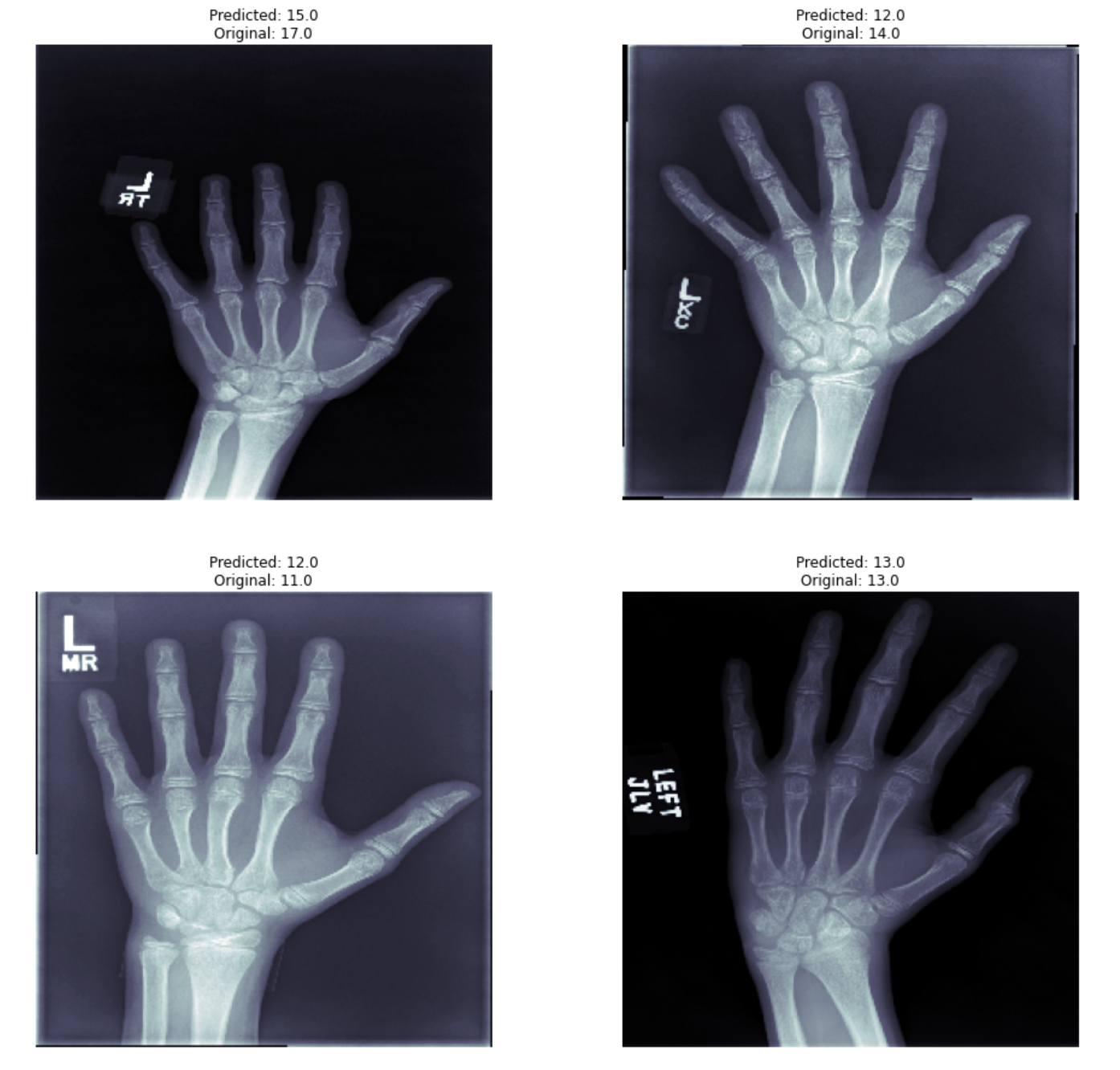}
	\caption{Output images with predicted ages by EfficientNetB4.} 
	\label{fig:4c}
\end{figure}

\begin{figure}
	\centering
	\includegraphics[width=0.5\textwidth]{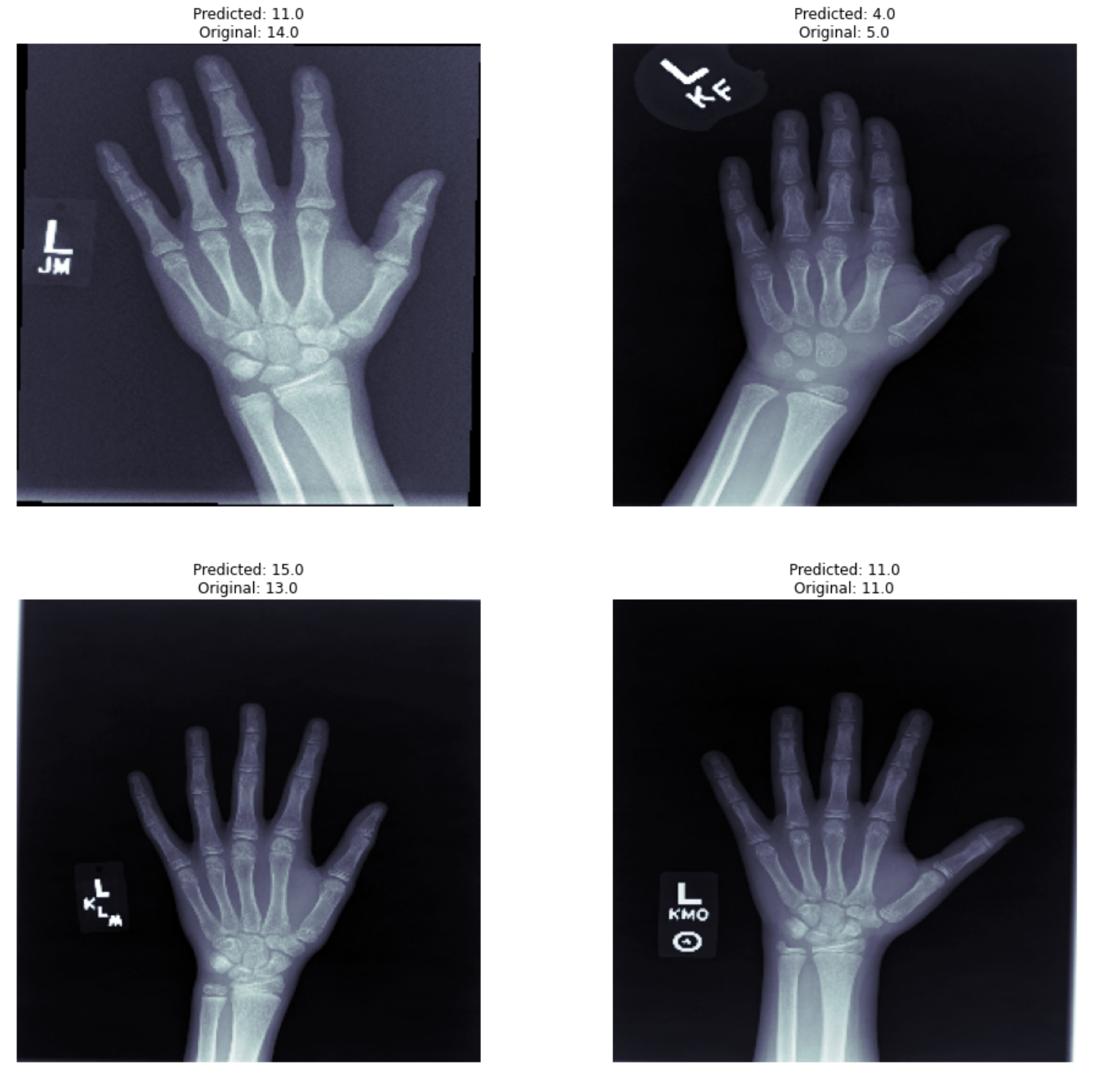}
	\caption{Output images with predicted ages by EN-AA.} 
	\label{fig:4d}
\end{figure}

\begin{figure}
	\centering
	\includegraphics[width=0.5\textwidth]{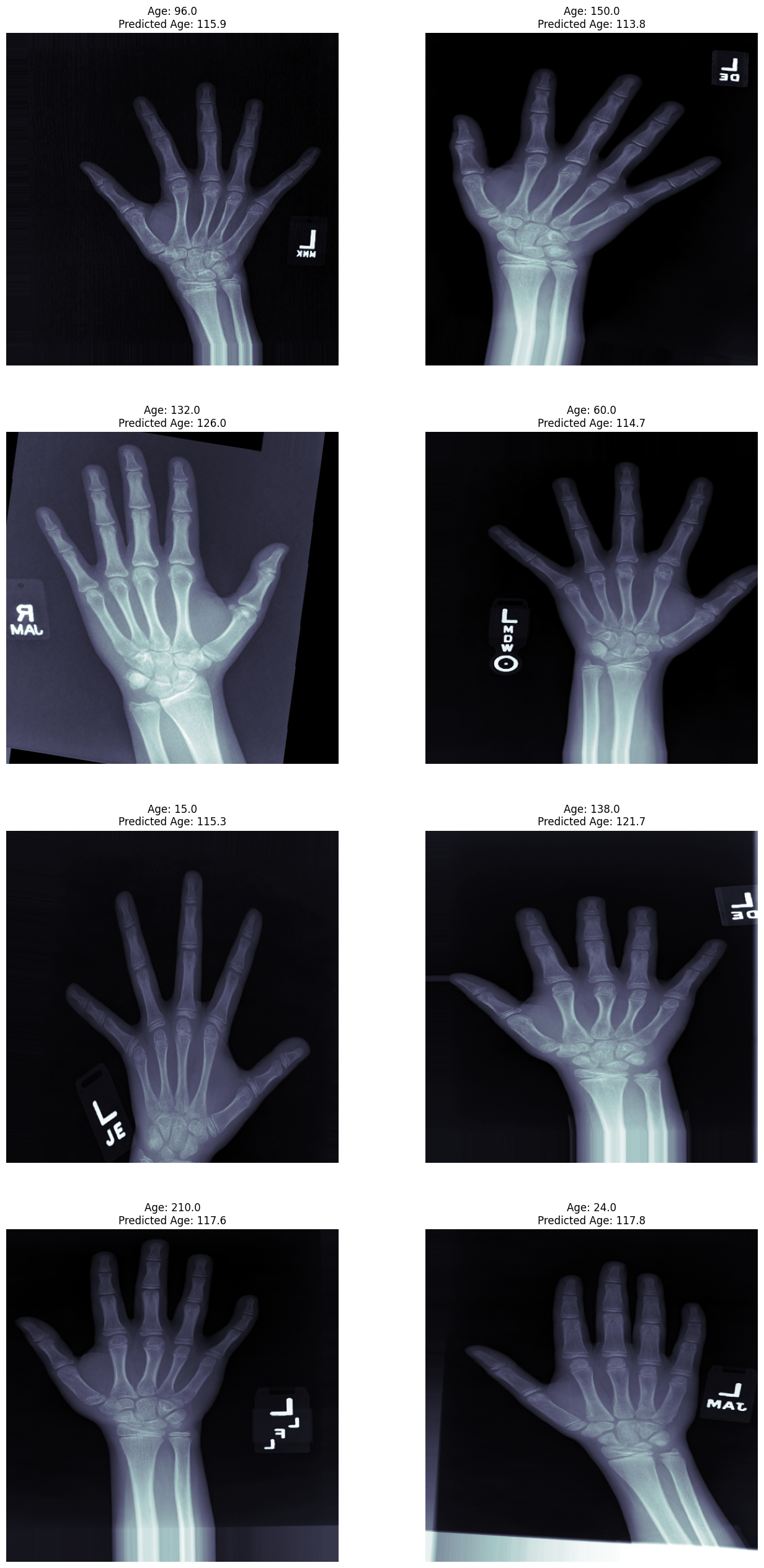}
	\caption{Output images with predicted ages by EfficientNetB0.} 
	\label{fig:4b}
\end{figure}

%results
The study focuses on predicting bone age from hand X-ray images of children using the EfficientNet variants and EN-AA model. The primary data source is the RSNA Bone Age Assessment dataset comprising hand X-ray images with corresponding bone ages. Employing transfer learning, the pre-trained EfficientNetB0 and EfficientNetB4 models were finetuned on this dataset. Data augmentation techniques like rotation, zooming, and flipping were utilized to enhance the model's efficacy, effectively expanding the dataset size. Further experimentation involved adjusting hyperparameters and refining data augmentation techniques to optimize model performance. The outcomes affirm the efficacy of using a pre-trained EfficientNet model coupled with data augmentation for bone age prediction.
The age predictions made by different EfficientNet models are illustrated in Figures \ref{fig:4a}, \ref{fig:4e}, and \ref{fig:4f}. These figures represent the results from EfficientNetB0, EfficientNetB4, and EN-AA, respectively. Additionally, Figures \ref{fig:4g} and \ref{fig:4h} display the learning curves for EfficientNetB4 and EN-AA. These demonstrate a decline in training and validation loss over epochs, indicating effective learning and model optimization.
A closer analysis of individual predictions reveals variations in performance. From Fig. \ref{fig:4c}, the predicted ages for EfficientNetB4 model can be observed with respect to the original age. For instance, EfficientNetB4 estimated the ages as 15, 12, 12, and 13 months against the original ages of 17, 14, 11, and 13 months, respectively. Similar outcomes can be observed for EN-AA model in Fig. \ref{fig:4d}. EN-AA predictions for ages 14, 5, 13, and 11 months were 11, 4, 15, and 11 months, respectively. Fig. \ref{fig:4b} represent the predicted ages by EfficientNetB0 model. The model predicted the ages as 115.9, 113.8, 126, 114.7, 115.3, 121.7, 117.6, and 117.8 months correspondingly for 96, 150, 132, 60, 15, 138, 210, and 24 months. These individual case studies highlight the nuanced performance differences between models. It clearly noticeable that EfficientNetB4 and EN-AA models were superior compared to the EfficientNetB0 model. Figures \ref{fig:4b}, \ref{fig:4c}, and \ref{fig:4d} further elucidate these differences by showcasing bone images with both predicted and original ages for EfficientNetB0, EfficientNetB4, and EN-AA models, respectively. These visual representations underscore the varying degrees of accuracy each model variant achieves. The study demonstrates the potential of using pre-trained EfficientNet models, enhanced by data augmentation strategies, for accurately predicting bone age from X-ray images. The comparative analysis of different EfficientNet models, including those enhanced with Additive Attention, provides valuable insights into the capabilities and limitations of these approaches in the context of bone age assessment.



\section{Conclusions}
\label{con}
In this study, we employed the EfficientNet models for the critical task of bone age estimation, which is integral to assessing human development and aiding in disease diagnosis. The process involved vital steps such as detailed data preprocessing: resizing, normalization, and enhancement to reduce data bias and expand the training set. EfficientNetB0 was pivotal in extracting intricate image features. These features, combined with gender data, were then processed through a fully connected (FC) layer, culminating in the bone age prediction. Using a pre-trained model was instrumental in achieving rapid convergence and enhancing the overall predictive accuracy. Additionally, the study delved into the capabilities of EfficientNetB4 and its variant with Additive Attention (EN-AA). Findings indicated that EfficientNetB4 outperformed the B0 model, with the Additive Attention variant demonstrating comparable proficiency. The study identifies two primary avenues for advancement: deploying advanced versions of EfficientNet coupled with specialized attention modules to more precisely target and analyze regions of interest in X-ray images and the extension of training epochs to refine the models further. These future endeavors aim to improve the accuracy of bone age predictions and prepare these models for broader clinical application and eventual market integration.

\bibliographystyle{IEEEtran}
\bibliography{mainRef}

\end{document}